\documentclass[
twocolumn
]{ceurart}

\usepackage{listings}
\usepackage{float}
\usepackage{amsmath}
\usepackage{hyperref}
\usepackage[noabbrev,capitalize,nameinlink]{cleveref}
\usepackage{xcolor, soul, colortbl}
\usepackage{array, multirow, graphicx}
\usepackage{adjustbox}
\usepackage{booktabs}
\usepackage{tikz}
\usepackage{tikz-dependency}
\usepackage{algorithm}
\usepackage{algpseudocode}
\newcommand*\circled[1]{\tikz[baseline=(char.base)]{
            \node[shape=circle,draw,inner sep=.6pt] (char) {#1};}}
            
\usepackage{microtype}
\definecolor{Gray}{gray}{0.9}
\definecolor{LightCyan}{rgb}{0.88,1,1}
\newcolumntype{?}{!{\vrule width 1pt}}
\newcolumntype{g}{>{\columncolor{LightCyan}}r}

\usepackage{amssymb}
\usepackage{pifont}
\newcommand{\say}{\textit{Sayfullina}}
\newcommand{\sks}{\textit{SkillSpan}}

\begin{document}

\copyrightyear{2022}
\copyrightclause{Copyright for this paper by its authors.
  Use permitted under Creative Commons License Attribution 4.0
  International (CC BY 4.0).}

\conference{RecSys in HR'22: The 2nd Workshop on Recommender Systems for Human Resources, in conjunction with the 16th ACM Conference on Recommender Systems, September 18--23, 2022, Seattle, USA.}

\title{Skill Extraction from Job Postings using Weak Supervision}


\author[1]{Mike Zhang}[%
email=mikz@itu.dk,
url=https://jjzha.github.io/,
]
\cormark[1]
\address[1]{IT University of Copenhagen, Rued Langgaards Vej 7, 2300, Copenhagen, Denmark}

\author[1]{Kristian N{\o}rgaard Jensen}[%
email=krnj@itu.dk,
url=http://kris927b.github.io/,
]

\author[1]{Rob {van der Goot}}[%
email=robv@itu.dk,
url=http://robvanderg.github.io/,
]

\author[1,2]{Barbara Plank}[%
email=b.plank@lmu.de,
url=http://bplank.github.io/,
]
\address[2]{Ludwig Maximilian University of Munich, Akademiestraße 7, 80799, Munich, Germany}

\cortext[1]{Corresponding author.}

\begin{abstract}
Aggregated data obtained from job postings provide powerful insights into labor market demands, and emerging skills, and aid job matching. However, most extraction approaches are supervised and thus need costly and time-consuming annotation. To overcome this, we propose Skill Extraction with Weak Supervision. We leverage the European Skills, Competences, Qualifications and Occupations taxonomy to find similar skills in job ads via latent representations. The method shows a strong positive signal, outperforming baselines based on token-level and syntactic patterns.
\end{abstract}


\maketitle

\section{Introduction}\label{intro}

The labor market is under constant development---often due to changes in technology, migration, and digitization---and so are the skill sets required~\cite{brynjolfsson2011race,brynjolfsson2014second}. Consequentially, large quantities of job vacancy data is emerging on a variety of platforms. Insights from this data on labor market skill set demands could aid, for instance, job matching~\cite{balog2012expertise}. The task of automatic \emph{skill extraction} (SE) is to extract the competences necessary for any occupation from unstructured text. 

Previous work on supervised SE frame it as a sequence labeling task (e.g.,~\cite{sayfullina2018learning,tamburri2020dataops,chernova2020occupational,zhang-jensen-plank:2022:LREC,zhang-etal-2022-skillspan,green-maynard-lin:2022:LREC,gnehm-bhlmann-clematide:2022:LREC}) or multi-label classification \cite{bhola-etal-2020-retrieving}. Annotation is a costly and time-consuming process with little annotation guidelines to work with. This could be alleviated by using predefined skill inventories. 

\begin{figure}
    \centering
    \includegraphics[width=.75\columnwidth]{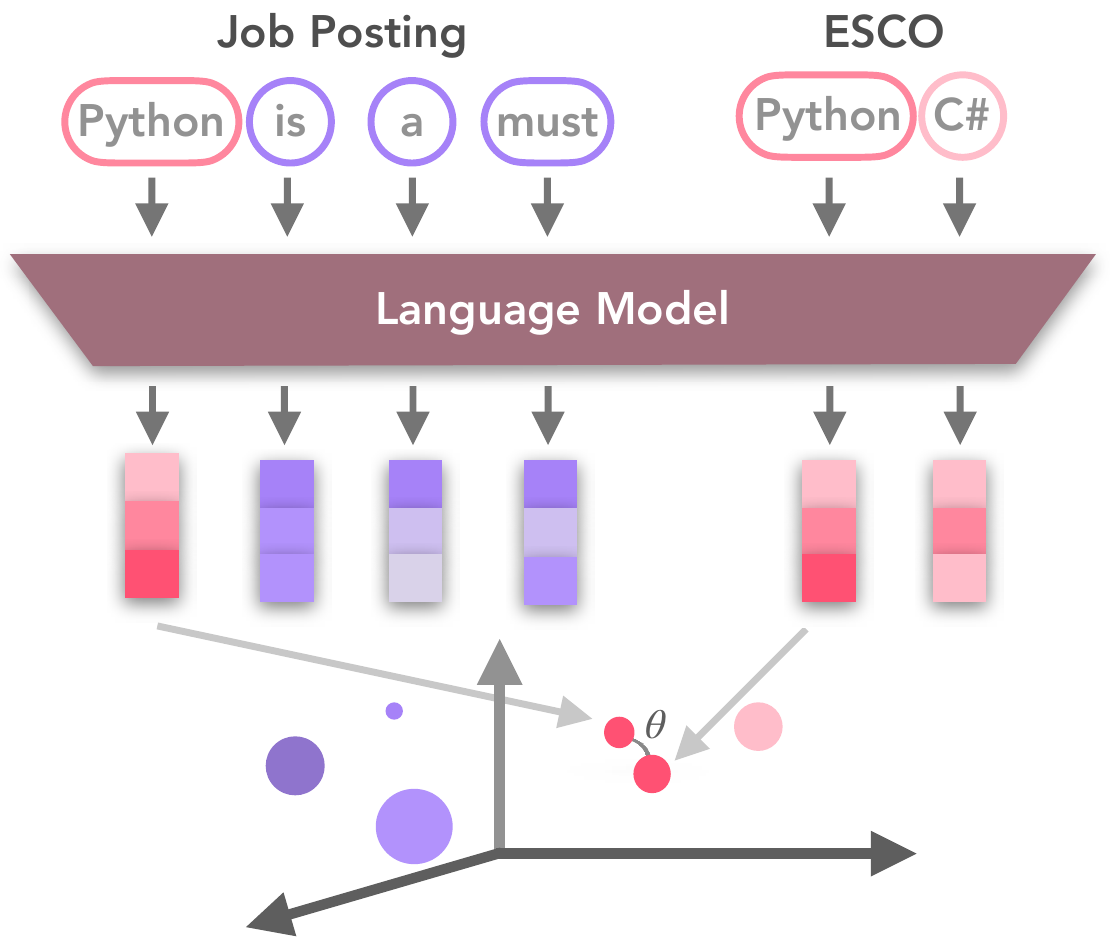}
    \caption{\textbf{Weakly Supervised Skill Extraction.} All ESCO skills and n-grams are extracted and embedded through a language model, e.g., RoBERTa~\cite{liu2019roberta}, to get representations. We label \emph{spans} from job postings close in vector space to the ESCO skill.}
    \looseness=-1
    \label{fig:fig1}
\end{figure}

In this work, we approach span-level SE with weak supervision: 
We leverage the European Skills, Competences, Qualifications and Occupations (ESCO;~\cite{le2014esco}) taxonomy and find similar spans that relate to ESCO skills in embedding space (\cref{fig:fig1}). The advantages are twofold: First, labeling skills becomes obsolete, which mitigates the cumbersome process of annotation. Second, by extracting skill phrases, this could possibly enrich skill inventories (e.g., ESCO) by finding paraphrases of existing skills. We seek to answer: \emph{How viable is Weak Supervision in the context of SE?} We contribute: \circled{1} A novel weakly supervised method for SE; \circled{2} A linguistic analysis of ESCO skills and their presence in job postings;
\circled{3} An empirical analysis of different embedding pooling methods for SE for two skill-based datasets.\footnote{\url{https://github.com/jjzha/skill-extraction-weak-supervision}}

\begin{figure*}[t]
    \centering
    \includegraphics[width=\linewidth]{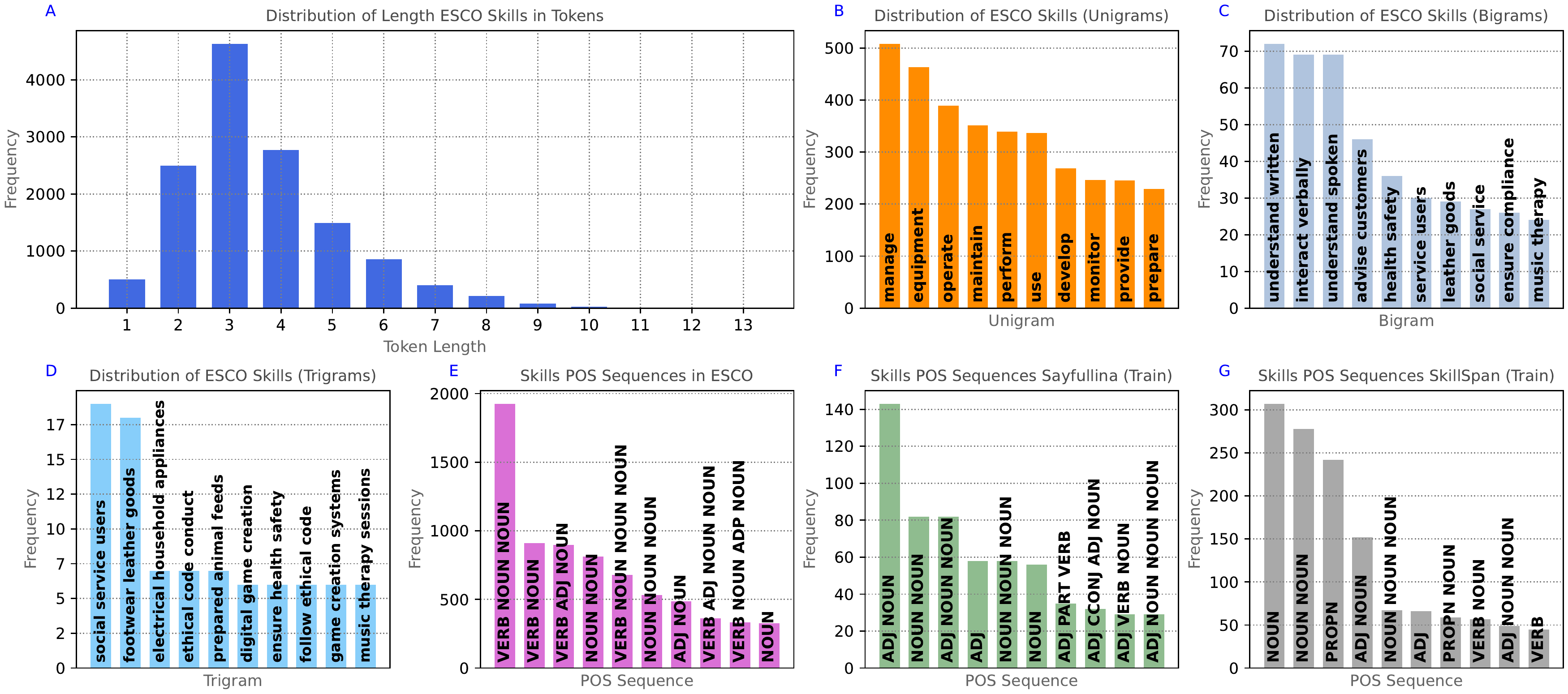} 
    \caption{\textbf{Surface-level Statistics of ESCO.} We show various statistics of ESCO. (\textbf{A}) ESCO skills token length, the mode is three tokens. (\textbf{B}) Most frequent unigrams of ESCO skills. (\textbf{C}) Most frequent bigrams of ESCO skills. (\textbf{D}) Most frequent trigrams of ESCO skills. (\textbf{E}) Most frequent POS sequences of ESCO skills. Last, we show the POS sequences of unique skills in both train sets of \say{} and \sks{} (\textbf{F}-\textbf{G}).}
    \label{fig:esco-stats}
\end{figure*}

\section{Methodology}
Formally, we consider a set of job postings $\mathcal{D}$, where $d \in \mathcal{D}$ is a set of sequences (e.g., job posting sentences) with the $i^\text{th}$ input sequence $\mathcal{T}^i_{d} = [t_1, t_2, ..., t_n]$ and a target sequence of \texttt{BIO}-labels $\mathcal{Y}^i_{d} = [y_1, y_2, ..., y_n]$ (e.g., ``\texttt{B-SKILL}'', ``\texttt{I-SKILL}'', ``\texttt{O}'').\footnote{Definition of labels can be found in~\cite{zhang-etal-2022-skillspan}.} The goal is to use an algorithm, which predicts skill spans by assigning an output label sequence $\mathcal{Y}^i_{d}$ for each token sequence $\mathcal{T}^i_{d}$ from a job posting based on representational similarity of a span to any skill in ESCO.

\begin{table}[t]
\centering
\caption{\textbf{Statistics of Datasets.} Indicated is each dataset and their respective number of sentences, tokens, skill spans, and the average length of skills in tokens.}
\resizebox{.88\linewidth}{!}{
\begin{tabular}{l|l|rr}
\toprule
& \textbf{Statistics} & \textbf{\say{}} & \textbf{\sks{}} \\
\midrule
\multirow{3}{*}{\rotatebox[origin=c]{90}{\textsc{\textbf{Train}}}}
  & \textbf{\# Sentences}              & 3,703     & 5,866     \\
  & \textbf{\# Tokens}                 & 53,095    & 122,608   \\
  & \textbf{\# Skill Spans}            & 3,703     & 3,325     \\
  \midrule
\multirow{3}{*}{\rotatebox[origin=c]{90}{\textsc{\textbf{Dev.}}}}
  & \textbf{\# Sentences}              & 1,856     & 3,992     \\
  & \textbf{\# Tokens}                 & 26,519    & 52,084     \\
  & \textbf{\# Skill Spans}            & 1,856     & 2,697     \\
  \midrule
\multirow{3}{*}{\rotatebox[origin=c]{90}{\textsc{\textbf{Test}}}}
  & \textbf{\# Sentences}              & 1,848     & 4,680     \\
  & \textbf{\# Tokens}                 & 26,569    & 57,528    \\
  & \textbf{\# Skill Spans}            & 1,848     & 3,093     \\
  \midrule
 & \textbf{Avg.\ Len.\ Skills }             & 1.77     & 2.92     \\
  \bottomrule
    \end{tabular}
    }
    \label{tab:num_post}
\end{table}

\subsection{Data}
We use the datasets from~\cite{zhang-etal-2022-skillspan} (\sks{}) and a modification of~\cite{sayfullina2018learning} (\say{}).\footnote{In contrast to \sks{}, \say{} has a skill in every sentence, where they focus on categorizing sentences for soft skills.}
In~\cref{tab:num_post}, we show the statistics of both. \sks{} contain nested labels for skill and knowledge components~\cite{le2014esco}. To make it fit for our weak supervision approach, we simplify their dataset by considering both skills and knowledge labels as one label (i.e., \texttt{B-KNOWLEDGE} becomes \texttt{B-SKILL}).

\paragraph{ESCO Statistics} 
We use ESCO as a weak supervision signal for discovering skills in job postings. There are 13,890 ESCO skills.\footnote{Per 25-03-2022, taking ESCO \texttt{v1.0.9}.} In~\cref{fig:esco-stats}, we show statistics of the taxonomy: (A) On average most skills are 3 tokens long. In (C-D), we show 
n-grams frequencies with range [$1;3$]. We can see that the most frequent uni- and bigrams are verbs, while the most frequent trigrams consist of nouns. 

Additionally, we show an analysis of ESCO skills from a linguistic perspective. We tag the training data using the publicly available MaChAmp v0.2 model~\cite{van-der-goot-etal-2021-massive} trained on all Universal Dependencies 2.7 treebanks~\cite{11234/1-3687}.\footnote{A Udify-based~\cite{kondratyuk201975} multi-task model for POS, lemmatization, dependency parsing, built on top of the \texttt{transformers} library~\cite{wolf-etal-2020-transformers}, and specifically using mBERT~\cite{devlin2019bert}.} Then, we count the most frequent Part-of-Speech (POS) tags in all sources of data (E-G). ESCO's most frequent tag sequences are \texttt{VERB-NOUN}, these are not as frequent in \say{} nor \sks{}. \say{} mostly consists of adjectives, which is attributed to the categorization of soft skills. \sks{} mostly consists of \texttt{NOUN} sequences. Overall, we observe most skills consist of verb and noun phrases.

\begin{figure*}[t]
    \centering
    \includegraphics[width=.95\linewidth]{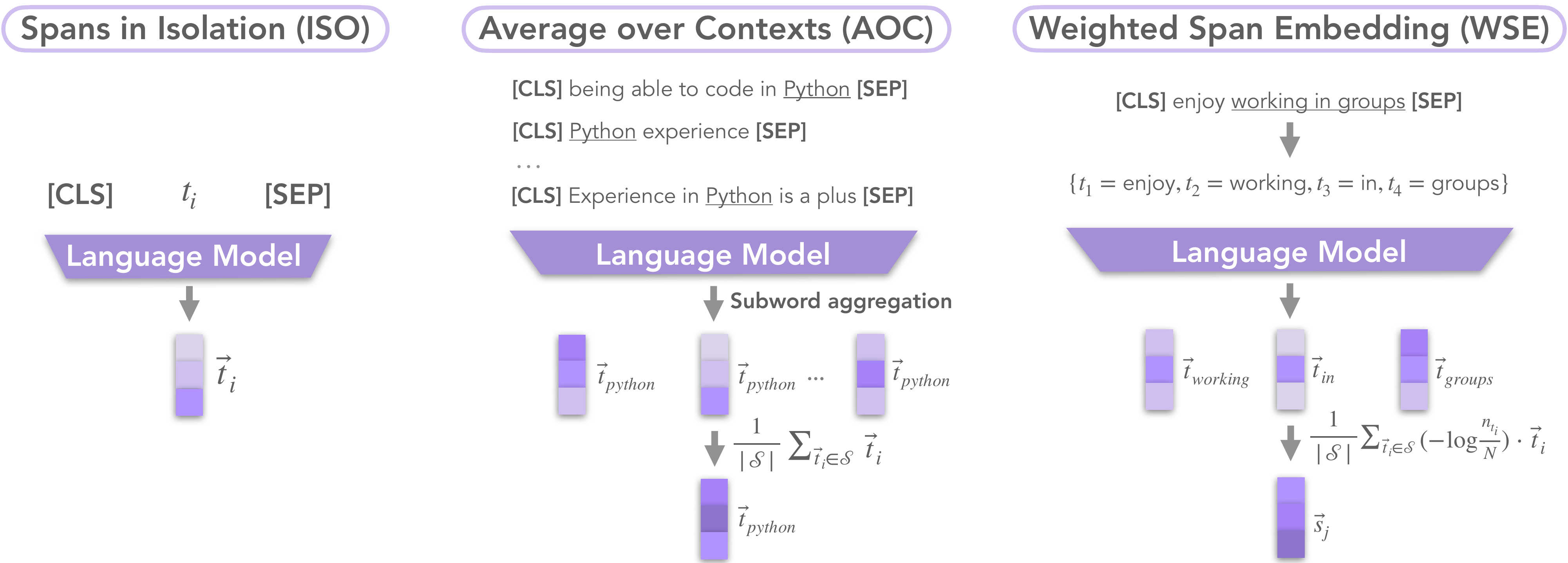}
    \caption{\textbf{Skill Representations.} We show different methods to embed ESCO skill phrases. The approaches are inspired by \protect\citet{litschko2022cross}. We embed a skill by encoding it directly without surrounding context (\textbf{left}). We aggregate different contextual representations of the same skill term (\textbf{middle}). Last, we encode the skill phrase via a weighted sum of embeddings with each token's inverse document frequency as weight (\textbf{right}). For the middle and right methods, $\mathcal{S}$ is the number of sentences where the ESCO skill appears.}
    \label{fig:vectors}
\end{figure*}

\subsection{Baselines}
As our approach is to find similar n-grams based on ESCO skills, we choose an n-gram range of $[1;4]$ (where $4$ is the median) derived from~\cref{fig:esco-stats} (A). For higher matching probability, we apply an additional pre-processing step to the ESCO skills by removing non-tokens (e.g., brackets) and words between brackets (e.g., ``Java (programming)'' becomes ``Java''). We have three baselines: 

\textbf{Exact Match}: We do exact substring matching with ESCO and the sentences in both datasets.

\textbf{Lemmatized Match}: ESCO skills are written in the infinitive form. We take the same approach as exact match on the training sets, now with the lemmatized data of both. The data is lemmatized with MaChAmp v0.2~\cite{van-der-goot-etal-2021-massive}.

\textbf{POS Sequence Match}: Motivated by the observation that certain POS sequences often overlap between sources (\cref{fig:esco-stats}, E-G), we attempt to match POS sequences within ESCO with the POS sequences in the datasets. For example \texttt{NOUN-NOUN}, \texttt{NOUN}, \texttt{VERB-NOUN} and \texttt{ADJ-NOUN} sequences are commonly occurring in all three sources.

\begin{algorithm}[t]
\caption{Weakly Supervised Skill Extraction}\label{alg:cap}
\begin{algorithmic}
\Require $M \in \{\text{RoBERTa}, \text{JobBERT}\}$
\Require $E \in \{\text{ISO},\text{AOC},\text{WSE}\}$
\Require $\tau \in [0,1]$
\State $P \gets D$ \Comment{A set of sentences from job postings}
\State $S \gets S_E$ \Comment{ESCO Skill embeddings of type $E$}
\State $L \gets \emptyset$
\For{$p \in P$}
    \State $\theta \gets 0$
    \For{$n \in p$} \Comment{Each ngram $n$ of size $1-4$}
        \State $E \gets M(n)$
        \State $\Theta \gets \text{CosSim}(S, E)$ 
        \If{$\text{max}(\Theta) > \tau \wedge \text{max}(\Theta) > \theta$}
            \State $\theta \gets \text{max}(\Theta)$
        \EndIf
    \EndFor
    \State $L \gets [L, \theta]$
\EndFor
\State \textbf{return} $L$
\end{algorithmic}
\end{algorithm}

\subsection{Skill Representations}\label{subsec:rep}

We investigate several encoding strategies to match n-gram representations to embedded ESCO skills, the approaches are inspired by~\citet{litschko2022cross}, where they applied them to Information Retrieval. The language models (LMs) used to encode the data are RoBERTa~\cite{liu2019roberta} and the domain-specific JobBERT~\cite{zhang-etal-2022-skillspan}. All obtained vector representations of skill phrases with the three previous encoding methods are compared pairwise with each n-gram created from \say{} and \sks{}. An explanation of the methods (see~\cref{fig:vectors}):

\textbf{Span in Isolation (ISO)}: We encode skill phrases $t$ from ESCO in isolation using the aforementioned LMs, without surrounding contexts.

\textbf{Average over Contexts (AOC)}: We leverage the surrounding context of a skill phrase $t$ by collecting all the sentences containing $t$. We use all available sentences in the job postings dataset (excluding \textsc{Test}). For a given job posting sentence, we encode $t$ by using one of the previous mentioned LMs. We average the embeddings of its constituent subwords to obtain the final embedding $t$.

\textbf{Weighted Span Embedding (WSE)}: We obtain all inverse document frequency (idf) values of each token $t_i$ via 

\begin{displaymath}
\text{idf} = -\text{log}\frac{n_{t_i}}{N},
\end{displaymath} 
where $n_{t_i}$ is the number of occurrences of $t_i$ and $N$ the total number of tokens in our dataset. We encode the input sentence and compute the weighted sum of the embeddings ($\vec{s}_j$) of the specific skill phrase in the sentence, where each $t_i$'s IDF scores are used as weights. Again, we only use the first subword token for each tokenized word. Formally, this is 

\begin{displaymath}
\vec{s}_j = \sum_{\vec{t}_i} (-\text{log}\frac{n_{t_i}}{N}) \cdot \vec{t}_i.
\end{displaymath}

\begin{figure*}[t]
    \centering
    \includegraphics[width=\linewidth]{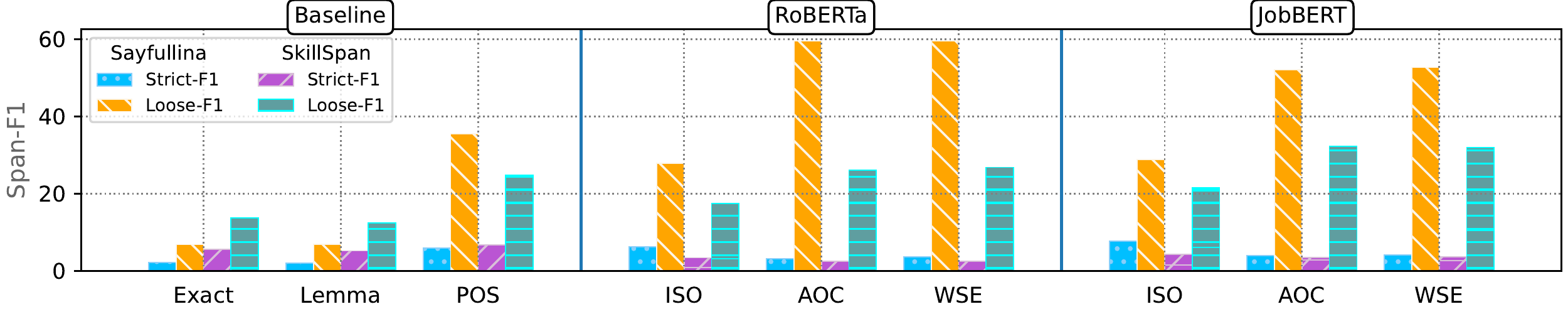}
    \looseness=-1
    \caption{\textbf{Results of Methods.} Results on \say{} and \sks{} are indicated by ``Baseline'' showing performance of Exact, Lemmatized (Lemma), and Part-of-Speech (POS). The performance of ISO, AOC, and WSE are separated by model, indicated by ``RoBERTa'' and ``JobBERT''. The performance of RoBERTa and JobBERT on \sks{} is determined by the best performing CosSim threshold (0.8).} 
    \label{fig:results-f1}
\end{figure*}

\begin{table*}[ht]
    \centering
    \caption{\textbf{Qualitative Examples of Predicted Spans.} We show the gold versus predicted spans of the best performing model on both datasets. The first 5 qualitative examples are from \say{} (RoBERTa with WSE), the last 5 are from \sks{}. Yellow the gold span and pink indicates the predicted span. The examples show many partial overlaps with the gold spans (but also incorrect ones), hence the high loose-F1.}
    \resizebox{\linewidth}{!}{
    \begin{tabular}{c|l|l}
    \toprule
    & \textsc{\textbf{Gold}} & \textsc{\textbf{Predicted}} \\
    \midrule
    \multirow{5}{*}{\rotatebox[origin=c]{90}{\textbf{\say{}}}}
    &...a \colorbox{yellow}{dynamic customer focused person} to join... & ...a dynamic \colorbox{pink}{customer focused person to} join... \\
    &...strong leadership and \colorbox{yellow}{team management skills}...                 & ...strong leadership \colorbox{pink}{and team management skills}...\\
    &...speak and \colorbox{yellow}{written english skills}...                          & ...speak and \colorbox{pink}{written english} skills...\\
    &...a \colorbox{yellow}{team environment and working independently skills}...                       & ...a team \colorbox{pink}{environment and working independently} skills...\\
    &...tangible business benefit extremely \colorbox{yellow}{articulate} and...        & ...tangible \colorbox{pink}{business benefit} extremely articulate and... \\
    \midrule
    \multirow{5}{*}{\rotatebox[origin=c]{90}{\textbf{\sks{}}}} 
    &...researcher within \colorbox{yellow}{machine learning} and \colorbox{yellow}{sensory system design}...     & ...researcher within machine \colorbox{pink}{learning and sensory system} design...\\
    &...standards and procedures \colorbox{yellow}{accessing and updating records}...   & ...standards and \colorbox{pink}{procedures accessing and updating} records...\\
    &...with a \colorbox{yellow}{passion for education} to... & ...with a passion for \colorbox{pink}{education} to...\\
    &...understands \colorbox{yellow}{Agile} as a mindset... & ...\colorbox{pink}{understands Agile as a} mindset...\\
    &...experience with \colorbox{yellow}{AWS} \colorbox{yellow}{GCP} \colorbox{yellow}{Microsoft Azure}... &  ...experience \colorbox{pink}{with AWS} GCP Microsoft...\\
    \bottomrule
    \end{tabular}}
    \label{tab:quali}
\end{table*}

\paragraph{Matching} 
We rank pairs of ESCO embeddings $\vec{t}$ and encoded candidate n-grams $\vec{g}$ in decreasing order of cosine similarity (CosSim), calculated as 

\begin{displaymath}
\text{CosSim(}\vec{t},\vec{g}\text{)} = \frac{\vec{t}^T \vec{g}}{\|\vec{t}\| \|\vec{g}\|}.
\end{displaymath}

We show our pseudocode of the matching algorithm in \cref{alg:cap}. Note that in \sks{} we have to set a threshold for CosSim, as there are sentences with no skills. A threshold allows us to have a ``no skill'' option. As seen in \cref{fig:results}, \cref{sec:appendix-exact-num} the threshold sensitivity on \sks{} differs for JobBERT: Performance fluctuates, compared to RoBERTa. Precision goes up with a higher threshold, while recall goes down. For RoBERTa, it stays similar until CosSim$=0.9$. We use CosSim$=0.8$  as over 2 LMs and 3 methods it provides the best cutoff.

\section{Analysis of Results}

\paragraph{Results}

Our main results (\cref{fig:results-f1}) show the baselines against ISO, AOC, and WSE of both datasets. We evaluate with two types of F1, following~\citet{van-der-goot-etal-2020-cross}: \texttt{strict} and \texttt{loose-F1}. For full model fine-tuning, RoBERTa achieves 91.31 and 98.55 strict and loose F1 on \say{} respectively. For \sks{}, this is 23.21 and 44.72 strict and loose F1 (on the available subsets of \sks{}). JobBERT achieves 90.18 and 98.19 strict and loose F1 on \say{}, 49.44 and 74.41 strict and loose F1 on \sks{}. The large difference between results is most likely due to lack of negatives in \say{}, i.e., all sentences contain a skill, which makes the task easier. These results highlight the difficulty of SE on \sks{}, where there are negatives as well (sentences with no skills).

The exact match baseline on \sks{} is higher than \say{}. We attribute this to \sks{} also containing ``hard skills'' (e.g., ``Python''), which is easier to match substrings with than ``soft skills''.\footnote{The exact numbers (+precision and recall) are in~\cref{tab:exact}, \cref{sec:appendix-exact-num}, including the definition of strict and loose-F1.}

For the performance of the skill representations on \say{}, RoBERTa and JobBERT outperform the Exact and Lemmatized baseline on strict-F1. For the POS baseline, only the ISO method of both models is slightly better. JobBERT performs better than RoBERTa in strict-F1 on both datasets. 

There is a substantial difference between strict and loose-F1 on both datasets. This indicates that there is partial overlap among the predicted and gold spans. RoBERTa performs best for \say{}, achieving $59.61$ loose-F1 with WSE. In addition, the best performing method for JobBERT is also WSE ($52.69$ loose-F1). For \sks{} we see a drop, JobBERT outperforms RoBERTa with AOC ($32.30$ vs.\ $26.10$ loose-F1) given a threshold of CosSim = 0.8. We hypothesize this drop in performance compared to \say{} could be attributed again to \sks{} containing negative examples as well (i.e., sentences with no skill).

\paragraph{Qualitative Analysis}
A qualitative analysis (\cref{tab:quali}) reveals there is strong partial overlap with gold vs.\ predicted spans on both datasets, e.g., ``...strong leadership and \emph{team management skills}...'' vs.\ ``...strong leadership \emph{and team management skills}...'', indicating the viability of this method.

\section{Conclusion}
We investigate whether the ESCO skill taxonomy suits as weak supervision signal for Skill Extraction. We apply several skill representation methods based on previous work. We show that using representations of ESCO skills can aid us in this task. We achieve high loose-F1, indicating there is partial overlap between the predicted and gold spans, but need refined off-set methods to get the correct span out (e.g., human post-editing or automatic methods such as candidate filtering). Nevertheless, we see this approach as a strong alternative for supervised Skill Extraction from job postings. 

Future work could include going towards multilingual Skill Extraction, as ESCO consists of 27 languages, exact matching should be trivial. For the other methods several considerations need to be taken into account, e.g., a POS-tagger and/or lemmatizer for another language and a language-specific model.

\begin{acknowledgments}
We thank the NLPnorth group for feedback on an earlier version of this paper---in particular, Elisa Bassignana and Max M\"uller-Eberstein for insightful discussions. We would also like to thank the anonymous reviewers for their comments to improve this paper. Last, we also thank NVIDIA and the ITU High-performance Computing cluster for computing resources. This research is supported by the Independent Research Fund Denmark (DFF) grant 9131-00019B.
\end{acknowledgments}

\bibliography{sample-ceur}

\begin{thebibliography}{20}
\expandafter\ifx\csname natexlab\endcsname\relax\def\natexlab#1{#1}\fi
\providecommand{\url}[1]{\texttt{#1}}
\providecommand{\href}[2]{#2}
\providecommand{\path}[1]{#1}
\providecommand{\DOIprefix}{doi:}
\providecommand{\ArXivprefix}{arXiv:}
\providecommand{\URLprefix}{URL: }
\providecommand{\Pubmedprefix}{pmid:}
\providecommand{\doi}[1]{\href{http://dx.doi.org/#1}{\path{#1}}}
\providecommand{\Pubmed}[1]{\href{pmid:#1}{\path{#1}}}
\providecommand{\bibinfo}[2]{#2}
\ifx\xfnm\relax \def\xfnm[#1]{\unskip,\space#1}\fi
\bibitem[{Brynjolfsson and McAfee(2011)}]{brynjolfsson2011race}
\bibinfo{author}{E.~Brynjolfsson}, \bibinfo{author}{A.~McAfee},
  \bibinfo{title}{Race against the machine: How the digital revolution is
  accelerating innovation, driving productivity, and irreversibly transforming
  employment and the economy}, \bibinfo{publisher}{Brynjolfsson and McAfee},
  \bibinfo{year}{2011}.
\bibitem[{Brynjolfsson and McAfee(2014)}]{brynjolfsson2014second}
\bibinfo{author}{E.~Brynjolfsson}, \bibinfo{author}{A.~McAfee},
  \bibinfo{title}{The second machine age: Work, progress, and prosperity in a
  time of brilliant technologies}, \bibinfo{publisher}{WW Norton \& Company},
  \bibinfo{year}{2014}.
\bibitem[{Balog et~al.(2012)Balog, Fang, De~Rijke, Serdyukov, and
  Si}]{balog2012expertise}
\bibinfo{author}{K.~Balog}, \bibinfo{author}{Y.~Fang},
  \bibinfo{author}{M.~De~Rijke}, \bibinfo{author}{P.~Serdyukov},
  \bibinfo{author}{L.~Si},
\newblock \bibinfo{title}{Expertise retrieval},
\newblock \bibinfo{journal}{Foundations and Trends in Information Retrieval}
  \bibinfo{volume}{6} (\bibinfo{year}{2012}) \bibinfo{pages}{127--256}.
\bibitem[{Sayfullina et~al.(2018)Sayfullina, Malmi, and
  Kannala}]{sayfullina2018learning}
\bibinfo{author}{L.~Sayfullina}, \bibinfo{author}{E.~Malmi},
  \bibinfo{author}{J.~Kannala},
\newblock \bibinfo{title}{Learning representations for soft skill matching},
\newblock in: \bibinfo{booktitle}{International Conference on Analysis of
  Images, Social Networks and Texts}, \bibinfo{year}{2018}, pp.
  \bibinfo{pages}{141--152}.
\bibitem[{Tamburri et~al.(2020)Tamburri, Van Den~Heuvel, and
  Garriga}]{tamburri2020dataops}
\bibinfo{author}{D.~A. Tamburri}, \bibinfo{author}{W.-J. Van Den~Heuvel},
  \bibinfo{author}{M.~Garriga},
\newblock \bibinfo{title}{Dataops for societal intelligence: a data pipeline
  for labor market skills extraction and matching},
\newblock in: \bibinfo{booktitle}{2020 IEEE 21st International Conference on
  Information Reuse and Integration for Data Science (IRI)},
  \bibinfo{organization}{IEEE}, \bibinfo{year}{2020}, pp.
  \bibinfo{pages}{391--394}.
\bibitem[{Chernova(2020)}]{chernova2020occupational}
\bibinfo{author}{M.~Chernova},
\newblock \bibinfo{title}{Occupational skills extraction with {FinBERT}},
\newblock \bibinfo{journal}{Master's Thesis}  (\bibinfo{year}{2020}).
\bibitem[{Zhang et~al.(2022{\natexlab{a}})Zhang, Jensen, and
  Plank}]{zhang-jensen-plank:2022:LREC}
\bibinfo{author}{M.~Zhang}, \bibinfo{author}{K.~N. Jensen},
  \bibinfo{author}{B.~Plank},
\newblock \bibinfo{title}{Kompetencer: Fine-grained skill classification in
  danish job postings via distant supervision and transfer learning},
\newblock in: \bibinfo{booktitle}{Proceedings of the Language Resources and
  Evaluation Conference}, \bibinfo{publisher}{European Language Resources
  Association}, \bibinfo{address}{Marseille, France},
  \bibinfo{year}{2022}{\natexlab{a}}, pp. \bibinfo{pages}{436--447}. \URLprefix
  \url{https://aclanthology.org/2022.lrec-1.46}.
\bibitem[{Zhang et~al.(2022{\natexlab{b}})Zhang, Jensen, Sonniks, and
  Plank}]{zhang-etal-2022-skillspan}
\bibinfo{author}{M.~Zhang}, \bibinfo{author}{K.~N. Jensen},
  \bibinfo{author}{S.~Sonniks}, \bibinfo{author}{B.~Plank},
\newblock \bibinfo{title}{{S}kill{S}pan: Hard and soft skill extraction from
  {E}nglish job postings},
\newblock in: \bibinfo{booktitle}{Proceedings of the 2022 Conference of the
  North American Chapter of the Association for Computational Linguistics:
  Human Language Technologies}, \bibinfo{publisher}{Association for
  Computational Linguistics}, \bibinfo{address}{Seattle, United States},
  \bibinfo{year}{2022}{\natexlab{b}}, pp. \bibinfo{pages}{4962--4984}.
\bibitem[{Green et~al.(2022)Green, Maynard, and
  Lin}]{green-maynard-lin:2022:LREC}
\bibinfo{author}{T.~Green}, \bibinfo{author}{D.~Maynard},
  \bibinfo{author}{C.~Lin},
\newblock \bibinfo{title}{Development of a benchmark corpus to support entity
  recognition in job descriptions},
\newblock in: \bibinfo{booktitle}{Proceedings of the Language Resources and
  Evaluation Conference}, \bibinfo{publisher}{European Language Resources
  Association}, \bibinfo{address}{Marseille, France}, \bibinfo{year}{2022}, pp.
  \bibinfo{pages}{1201--1208}. \URLprefix
  \url{https://aclanthology.org/2022.lrec-1.128}.
\bibitem[{Gnehm et~al.(2022)Gnehm, BÃ¼hlmann, and
  Clematide}]{gnehm-bhlmann-clematide:2022:LREC}
\bibinfo{author}{A.-S. Gnehm}, \bibinfo{author}{E.~BÃ¼hlmann},
  \bibinfo{author}{S.~Clematide},
\newblock \bibinfo{title}{Evaluation of transfer learning and domain adaptation
  for analyzing german-speaking job advertisements},
\newblock in: \bibinfo{booktitle}{Proceedings of the Language Resources and
  Evaluation Conference}, \bibinfo{publisher}{European Language Resources
  Association}, \bibinfo{address}{Marseille, France}, \bibinfo{year}{2022}, pp.
  \bibinfo{pages}{3892--3901}. \URLprefix
  \url{https://aclanthology.org/2022.lrec-1.414}.
\bibitem[{Bhola et~al.(2020)Bhola, Halder, Prasad, and
  Kan}]{bhola-etal-2020-retrieving}
\bibinfo{author}{A.~Bhola}, \bibinfo{author}{K.~Halder},
  \bibinfo{author}{A.~Prasad}, \bibinfo{author}{M.-Y. Kan},
\newblock \bibinfo{title}{Retrieving skills from job descriptions: A language
  model based extreme multi-label classification framework},
\newblock in: \bibinfo{booktitle}{Proceedings of the 28th International
  Conference on Computational Linguistics}, \bibinfo{publisher}{International
  Committee on Computational Linguistics}, \bibinfo{address}{Barcelona, Spain
  (Online)}, \bibinfo{year}{2020}, pp. \bibinfo{pages}{5832--5842}.
\bibitem[{le~Vrang et~al.(2014)le~Vrang, Papantoniou, Pauwels, Fannes,
  Vandensteen, and De~Smedt}]{le2014esco}
\bibinfo{author}{M.~le~Vrang}, \bibinfo{author}{A.~Papantoniou},
  \bibinfo{author}{E.~Pauwels}, \bibinfo{author}{P.~Fannes},
  \bibinfo{author}{D.~Vandensteen}, \bibinfo{author}{J.~De~Smedt},
\newblock \bibinfo{title}{Esco: Boosting job matching in europe with semantic
  interoperability},
\newblock \bibinfo{journal}{Computer} \bibinfo{volume}{47}
  (\bibinfo{year}{2014}) \bibinfo{pages}{57--64}.
\bibitem[{Liu et~al.(2019)Liu, Ott, Goyal, Du, Joshi, Chen, Levy, Lewis,
  Zettlemoyer, and Stoyanov}]{liu2019roberta}
\bibinfo{author}{Y.~Liu}, \bibinfo{author}{M.~Ott}, \bibinfo{author}{N.~Goyal},
  \bibinfo{author}{J.~Du}, \bibinfo{author}{M.~Joshi},
  \bibinfo{author}{D.~Chen}, \bibinfo{author}{O.~Levy},
  \bibinfo{author}{M.~Lewis}, \bibinfo{author}{L.~Zettlemoyer},
  \bibinfo{author}{V.~Stoyanov},
\newblock \bibinfo{title}{Roberta: A robustly optimized bert pretraining
  approach},
\newblock \bibinfo{journal}{arXiv preprint arXiv:1907.11692}
  (\bibinfo{year}{2019}).
\bibitem[{van~der Goot et~al.(2021)van~der Goot, {\"U}st{\"u}n, Ramponi,
  Sharaf, and Plank}]{van-der-goot-etal-2021-massive}
\bibinfo{author}{R.~van~der Goot}, \bibinfo{author}{A.~{\"U}st{\"u}n},
  \bibinfo{author}{A.~Ramponi}, \bibinfo{author}{I.~Sharaf},
  \bibinfo{author}{B.~Plank},
\newblock \bibinfo{title}{Massive choice, ample tasks ({M}a{C}h{A}mp): A
  toolkit for multi-task learning in {NLP}},
\newblock in: \bibinfo{booktitle}{Proceedings of the 16th Conference of the
  European Chapter of the Association for Computational Linguistics: System
  Demonstrations}, \bibinfo{publisher}{Association for Computational
  Linguistics}, \bibinfo{address}{Online}, \bibinfo{year}{2021}, pp.
  \bibinfo{pages}{176--197}.
\bibitem[{Zeman et~al.(2021)Zeman, Nivre, Abrams, Ackermann, Aepli, Aghaei,
  Agi{\'c}, Ahmadi, Ahrenberg, Ajede, Aleksandravi{\v c}i{\=u}t{\.e}, Alfina,
  Antonsen, Aplonova, Aquino, Aragon, Aranzabe, Ar{\i}can, Arnard{\'o}ttir,
  Arutie, Arwidarasti, Asahara, Aslan, Ateyah, Atmaca, Attia, Atutxa,
  Augustinus, Badmaeva, Balasubramani, Ballesteros, Banerjee, Bank,
  Barbu~Mititelu, Barkarson, Basmov, Batchelor, Bauer, Bedir, Bengoetxea, Berk,
  Berzak, Bhat, Bhat, Biagetti, Bick, Bielinskien{\.e}, Bjarnad{\'o}ttir,
  Blokland, Bobicev, Boizou, Borges~V{\"o}lker, B{\"o}rstell, Bosco, Bouma,
  Bowman, Boyd, Braggaar, Brokait{\.e}, Burchardt, Candito, Caron, Caron,
  Cassidy, Cavalcanti, Cebiro{\u g}lu~Eryi{\u g}it, Cecchini, Celano, {\v
  C}{\'e}pl{\"o}, Cesur, Cetin, {\c C}etino{\u g}lu, Chalub, Chauhan, Chi,
  Chika, Cho, Choi, Chun, Cignarella, Cinkov{\'a}, Collomb, {\c C}{\"o}ltekin,
  Connor, Courtin, Cristescu, Daniel, Davidson, de~Marneffe, de~Paiva, Derin,
  de~Souza, Diaz~de Ilarraza, Dickerson, Dinakaramani, Di~Nuovo, Dione, Dirix,
  Dobrovoljc, Dozat, Droganova, Dwivedi, Eckhoff, Eiche, Eli, Elkahky, Ephrem,
  Erina, Erjavec, Etienne, Evelyn, Facundes, Farkas, Fernanda,
  Fernandez~Alcalde, Foster, Freitas, Fujita, Gajdo{\v s}ov{\'a}, Galbraith,
  Garcia, G{\"a}rdenfors, Garza, Gerardi, Gerdes, Ginter, Godoy, Goenaga,
  Gojenola, G{\"o}k{\i}rmak, Goldberg, G{\'o}mez~Guinovart,
  Gonz{\'a}lez~Saavedra, Grici{\=u}t{\.e}, Grioni, Grobol, Gr{\=
  u}z{\={\i}}tis, Guillaume, Guillot-Barbance, G{\"u}ng{\"o}r, Habash,
  Hafsteinsson, Haji{\v c}, Haji{\v c}~jr., H{\"a}m{\"a}l{\"a}inen,
  H{\`a}~M{\~y}, Han, Hanifmuti, Hardwick, Harris, Haug, Heinecke, Hellwig,
  Hennig, Hladk{\'a}, Hlav{\'a}{\v c}ov{\'a}, Hociung, Hohle, Huber, Hwang,
  Ikeda, Ingason, Ion, Irimia, Ishola, Ito, Jel{\'{\i}}nek, Jha, Johannsen,
  J{\'o}nsd{\'o}ttir, J{\o}rgensen, Juutinen, K, Ka{\c s}{\i}kara, Kaasen,
  Kabaeva, Kahane, Kanayama, Kanerva, Kara, Katz, Kayadelen, Kenney,
  Kettnerov{\'a}, Kirchner, Klementieva, K{\"o}hn, K{\"o}ksal, Kopacewicz,
  Korkiakangas, Kotsyba, Kovalevskait{\.e}, Krek, Krishnamurthy, Kuyruk{\c c}u,
  Kuzgun, Kwak, Laippala, Lam, Lambertino, Lando, Larasati, Lavrentiev, Lee,
  L{\^e}~H{\`{\^o}}ng, Lenci, Lertpradit, Leung, Levina, Li, Li, Li, Li, Lim,
  Lima~Padovani, Lind{\'e}n, Ljube{\v s}i{\'c}, Loginova, Luthfi, Luukko,
  Lyashevskaya, Lynn, Macketanz, Makazhanov, Mandl, Manning, Manurung, Mar{\c
  s}an, M{\u a}r{\u a}nduc, Mare{\v c}ek, Marheinecke, Mart{\'{\i}}nez~Alonso,
  Martins, Ma{\v s}ek, Matsuda, Matsumoto, Mazzei, {McDonald}, {McGuinness},
  Mendon{\c c}a, Miekka, Mischenkova, Misirpashayeva, Missil{\"a}, Mititelu,
  Mitrofan, Miyao, Mojiri~Foroushani, Moln{\'a}r, Moloodi, Montemagni, More,
  Moreno~Romero, Moretti, Mori, Mori, Morioka, Moro, Mortensen, Moskalevskyi,
  Muischnek, Munro, Murawaki, M{\"u}{\"u}risep, Nainwani, Nakhl{\'e},
  Navarro~Hor{\~n}iacek, Nedoluzhko, Ne{\v s}pore-B{\=e}rzkalne, Nevaci,
  Nguy{\~{\^e}}n~Th{\d i}, Nguy{\~{\^e}}n Th{\d i}~Minh, Nikaido, Nikolaev,
  Nitisaroj, Nourian, Nurmi, Ojala, Ojha, Ol{\'u}{\`o}kun, Omura, Onwuegbuzia,
  Osenova, {\"O}stling, {\O}vrelid, {\"O}zate{\c s}, {\"O}z{\c c}elik,
  {\"O}zg{\"u}r, {\"O}zt{\"u}rk~Ba{\c s}aran, Park, Partanen, Pascual,
  Passarotti, Patejuk, Paulino-Passos, Peljak-{\L}api{\'n}ska, Peng, Perez,
  Perkova, Perrier, Petrov, Petrova, Phelan, Piitulainen, Pirinen, Pitler,
  Plank, Poibeau, Ponomareva, Popel, Pretkalni{\c n}a, Pr{\'e}vost, Prokopidis,
  Przepi{\'o}rkowski, Puolakainen, Pyysalo, Qi, R{\"a}{\"a}bis, Rademaker,
  Rama, Ramasamy, Ramisch, Rashel, Rasooli, Ravishankar, Real, Rebeja, Reddy,
  Rehm, Riabov, Rie{\ss}ler, Rimkut{\.e}, Rinaldi, Rituma, Rocha,
  R{\"o}gnvaldsson, Romanenko, Rosa, Roșca, Rovati, Rudina, Rueter,
  R{\'u}narsson, Sadde, Safari, Sagot, Sahala, Saleh, Salomoni, Samard{\v
  z}i{\'c}, Samson, Sanguinetti, San{\i}yar, S{\"a}rg, Saul{\={\i}}te,
  Sawanakunanon, Saxena, Scannell, Scarlata, Schneider, Schuster, Schwartz,
  Seddah, Seeker, Seraji, Shen, Shimada, Shirasu, Shishkina, Shohibussirri,
  Sichinava, Siewert, Sigurðsson, Silveira, Silveira, Simi, Simionescu,
  Simk{\'o}, {\v S}imkov{\'a}, Simov, Skachedubova, Smith, Soares-Bastos,
  Spadine, Sprugnoli, Steingr{\'{\i}}msson, Stella, Straka, Strickland,
  Strnadov{\'a}, Suhr, Sulestio, Sulubacak, Suzuki, Sz{\'a}nt{\'o}, Taji,
  Takahashi, Tamburini, Tan, Tanaka, Tella, Tellier, Testori, Thomas, Torga,
  Toska, Trosterud, Trukhina, Tsarfaty, T{\"u}rk, Tyers, Uematsu, Untilov,
  Ure{\v s}ov{\'a}, Uria, Uszkoreit, Utka, Vajjala, van~der Goot, Vanhove, van
  Niekerk, van Noord, Varga, Villemonte de~la Clergerie, Vincze, Vlasova,
  Wakasa, Wallenberg, Wallin, Walsh, Wang, Washington, Wendt, Widmer, Williams,
  Wir{\'e}n, Wittern, Woldemariam, Wong, Wr{\'o}blewska, Yako, Yamashita,
  Yamazaki, Yan, Yasuoka, Yavrumyan, Yenice, Y{\i}ld{\i}z, Yu, {\v
  Z}abokrtsk{\'y}, Zahra, Zeldes, Zhu, Zhuravleva, and Ziane}]{11234/1-3687}
\bibinfo{author}{D.~Zeman}, \bibinfo{author}{J.~Nivre},
  \bibinfo{author}{M.~Abrams}, \bibinfo{author}{E.~Ackermann},
  \bibinfo{author}{N.~Aepli}, \bibinfo{author}{H.~Aghaei}, \bibinfo{author}{{\v
  Z}.~Agi{\'c}}, \bibinfo{author}{A.~Ahmadi}, \bibinfo{author}{L.~Ahrenberg},
  \bibinfo{author}{C.~K. Ajede}, \bibinfo{author}{G.~Aleksandravi{\v
  c}i{\=u}t{\.e}}, \bibinfo{author}{I.~Alfina}, \bibinfo{author}{L.~Antonsen},
  \bibinfo{author}{K.~Aplonova}, \bibinfo{author}{A.~Aquino},
  \bibinfo{author}{C.~Aragon}, \bibinfo{author}{M.~J. Aranzabe},
  \bibinfo{author}{B.~N. Ar{\i}can}, \bibinfo{author}{{\t H}.~Arnard{\'o}ttir},
  \bibinfo{author}{G.~Arutie}, \bibinfo{author}{J.~N. Arwidarasti},
  \bibinfo{author}{M.~Asahara}, \bibinfo{author}{D.~B. Aslan},
  \bibinfo{author}{L.~Ateyah}, \bibinfo{author}{F.~Atmaca},
  \bibinfo{author}{M.~Attia}, \bibinfo{author}{A.~Atutxa},
  \bibinfo{author}{L.~Augustinus}, \bibinfo{author}{E.~Badmaeva},
  \bibinfo{author}{K.~Balasubramani}, \bibinfo{author}{M.~Ballesteros},
  \bibinfo{author}{E.~Banerjee}, \bibinfo{author}{S.~Bank},
  \bibinfo{author}{V.~Barbu~Mititelu}, \bibinfo{author}{S.~Barkarson},
  \bibinfo{author}{V.~Basmov}, \bibinfo{author}{C.~Batchelor},
  \bibinfo{author}{J.~Bauer}, \bibinfo{author}{S.~T. Bedir},
  \bibinfo{author}{K.~Bengoetxea}, \bibinfo{author}{G.~Berk},
  \bibinfo{author}{Y.~Berzak}, \bibinfo{author}{I.~A. Bhat},
  \bibinfo{author}{R.~A. Bhat}, \bibinfo{author}{E.~Biagetti},
  \bibinfo{author}{E.~Bick}, \bibinfo{author}{A.~Bielinskien{\.e}},
  \bibinfo{author}{K.~Bjarnad{\'o}ttir}, \bibinfo{author}{R.~Blokland},
  \bibinfo{author}{V.~Bobicev}, \bibinfo{author}{L.~Boizou},
  \bibinfo{author}{E.~Borges~V{\"o}lker}, \bibinfo{author}{C.~B{\"o}rstell},
  \bibinfo{author}{C.~Bosco}, \bibinfo{author}{G.~Bouma},
  \bibinfo{author}{S.~Bowman}, \bibinfo{author}{A.~Boyd},
  \bibinfo{author}{A.~Braggaar}, \bibinfo{author}{K.~Brokait{\.e}},
  \bibinfo{author}{A.~Burchardt}, \bibinfo{author}{M.~Candito},
  \bibinfo{author}{B.~Caron}, \bibinfo{author}{G.~Caron},
  \bibinfo{author}{L.~Cassidy}, \bibinfo{author}{T.~Cavalcanti},
  \bibinfo{author}{G.~Cebiro{\u g}lu~Eryi{\u g}it}, \bibinfo{author}{F.~M.
  Cecchini}, \bibinfo{author}{G.~G.~A. Celano}, \bibinfo{author}{S.~{\v
  C}{\'e}pl{\"o}}, \bibinfo{author}{N.~Cesur}, \bibinfo{author}{S.~Cetin},
  \bibinfo{author}{{\"O}.~{\c C}etino{\u g}lu}, \bibinfo{author}{F.~Chalub},
  \bibinfo{author}{S.~Chauhan}, \bibinfo{author}{E.~Chi},
  \bibinfo{author}{T.~Chika}, \bibinfo{author}{Y.~Cho},
  \bibinfo{author}{J.~Choi}, \bibinfo{author}{J.~Chun}, \bibinfo{author}{A.~T.
  Cignarella}, \bibinfo{author}{S.~Cinkov{\'a}}, \bibinfo{author}{A.~Collomb},
  \bibinfo{author}{{\c C}.~{\c C}{\"o}ltekin}, \bibinfo{author}{M.~Connor},
  \bibinfo{author}{M.~Courtin}, \bibinfo{author}{M.~Cristescu},
  \bibinfo{author}{P.~Daniel}, \bibinfo{author}{E.~Davidson},
  \bibinfo{author}{M.-C. de~Marneffe}, \bibinfo{author}{V.~de~Paiva},
  \bibinfo{author}{M.~O. Derin}, \bibinfo{author}{E.~de~Souza},
  \bibinfo{author}{A.~Diaz~de Ilarraza}, \bibinfo{author}{C.~Dickerson},
  \bibinfo{author}{A.~Dinakaramani}, \bibinfo{author}{E.~Di~Nuovo},
  \bibinfo{author}{B.~Dione}, \bibinfo{author}{P.~Dirix},
  \bibinfo{author}{K.~Dobrovoljc}, \bibinfo{author}{T.~Dozat},
  \bibinfo{author}{K.~Droganova}, \bibinfo{author}{P.~Dwivedi},
  \bibinfo{author}{H.~Eckhoff}, \bibinfo{author}{S.~Eiche},
  \bibinfo{author}{M.~Eli}, \bibinfo{author}{A.~Elkahky},
  \bibinfo{author}{B.~Ephrem}, \bibinfo{author}{O.~Erina},
  \bibinfo{author}{T.~Erjavec}, \bibinfo{author}{A.~Etienne},
  \bibinfo{author}{W.~Evelyn}, \bibinfo{author}{S.~Facundes},
  \bibinfo{author}{R.~Farkas}, \bibinfo{author}{M.~Fernanda},
  \bibinfo{author}{H.~Fernandez~Alcalde}, \bibinfo{author}{J.~Foster},
  \bibinfo{author}{C.~Freitas}, \bibinfo{author}{K.~Fujita},
  \bibinfo{author}{K.~Gajdo{\v s}ov{\'a}}, \bibinfo{author}{D.~Galbraith},
  \bibinfo{author}{M.~Garcia}, \bibinfo{author}{M.~G{\"a}rdenfors},
  \bibinfo{author}{S.~Garza}, \bibinfo{author}{F.~F. Gerardi},
  \bibinfo{author}{K.~Gerdes}, \bibinfo{author}{F.~Ginter},
  \bibinfo{author}{G.~Godoy}, \bibinfo{author}{I.~Goenaga},
  \bibinfo{author}{K.~Gojenola}, \bibinfo{author}{M.~G{\"o}k{\i}rmak},
  \bibinfo{author}{Y.~Goldberg}, \bibinfo{author}{X.~G{\'o}mez~Guinovart},
  \bibinfo{author}{B.~Gonz{\'a}lez~Saavedra},
  \bibinfo{author}{B.~Grici{\=u}t{\.e}}, \bibinfo{author}{M.~Grioni},
  \bibinfo{author}{L.~Grobol}, \bibinfo{author}{N.~Gr{\= u}z{\={\i}}tis},
  \bibinfo{author}{B.~Guillaume}, \bibinfo{author}{C.~Guillot-Barbance},
  \bibinfo{author}{T.~G{\"u}ng{\"o}r}, \bibinfo{author}{N.~Habash},
  \bibinfo{author}{H.~Hafsteinsson}, \bibinfo{author}{J.~Haji{\v c}},
  \bibinfo{author}{J.~Haji{\v c}~jr.},
  \bibinfo{author}{M.~H{\"a}m{\"a}l{\"a}inen},
  \bibinfo{author}{L.~H{\`a}~M{\~y}}, \bibinfo{author}{N.-R. Han},
  \bibinfo{author}{M.~Y. Hanifmuti}, \bibinfo{author}{S.~Hardwick},
  \bibinfo{author}{K.~Harris}, \bibinfo{author}{D.~Haug},
  \bibinfo{author}{J.~Heinecke}, \bibinfo{author}{O.~Hellwig},
  \bibinfo{author}{F.~Hennig}, \bibinfo{author}{B.~Hladk{\'a}},
  \bibinfo{author}{J.~Hlav{\'a}{\v c}ov{\'a}}, \bibinfo{author}{F.~Hociung},
  \bibinfo{author}{P.~Hohle}, \bibinfo{author}{E.~Huber},
  \bibinfo{author}{J.~Hwang}, \bibinfo{author}{T.~Ikeda},
  \bibinfo{author}{A.~K. Ingason}, \bibinfo{author}{R.~Ion},
  \bibinfo{author}{E.~Irimia}, \bibinfo{author}{{\d O}.~Ishola},
  \bibinfo{author}{K.~Ito}, \bibinfo{author}{T.~Jel{\'{\i}}nek},
  \bibinfo{author}{A.~Jha}, \bibinfo{author}{A.~Johannsen},
  \bibinfo{author}{H.~J{\'o}nsd{\'o}ttir}, \bibinfo{author}{F.~J{\o}rgensen},
  \bibinfo{author}{M.~Juutinen}, \bibinfo{author}{S.~K},
  \bibinfo{author}{H.~Ka{\c s}{\i}kara}, \bibinfo{author}{A.~Kaasen},
  \bibinfo{author}{N.~Kabaeva}, \bibinfo{author}{S.~Kahane},
  \bibinfo{author}{H.~Kanayama}, \bibinfo{author}{J.~Kanerva},
  \bibinfo{author}{N.~Kara}, \bibinfo{author}{B.~Katz},
  \bibinfo{author}{T.~Kayadelen}, \bibinfo{author}{J.~Kenney},
  \bibinfo{author}{V.~Kettnerov{\'a}}, \bibinfo{author}{J.~Kirchner},
  \bibinfo{author}{E.~Klementieva}, \bibinfo{author}{A.~K{\"o}hn},
  \bibinfo{author}{A.~K{\"o}ksal}, \bibinfo{author}{K.~Kopacewicz},
  \bibinfo{author}{T.~Korkiakangas}, \bibinfo{author}{N.~Kotsyba},
  \bibinfo{author}{J.~Kovalevskait{\.e}}, \bibinfo{author}{S.~Krek},
  \bibinfo{author}{P.~Krishnamurthy}, \bibinfo{author}{O.~Kuyruk{\c c}u},
  \bibinfo{author}{A.~Kuzgun}, \bibinfo{author}{S.~Kwak},
  \bibinfo{author}{V.~Laippala}, \bibinfo{author}{L.~Lam},
  \bibinfo{author}{L.~Lambertino}, \bibinfo{author}{T.~Lando},
  \bibinfo{author}{S.~D. Larasati}, \bibinfo{author}{A.~Lavrentiev},
  \bibinfo{author}{J.~Lee}, \bibinfo{author}{P.~L{\^e}~H{\`{\^o}}ng},
  \bibinfo{author}{A.~Lenci}, \bibinfo{author}{S.~Lertpradit},
  \bibinfo{author}{H.~Leung}, \bibinfo{author}{M.~Levina},
  \bibinfo{author}{C.~Y. Li}, \bibinfo{author}{J.~Li}, \bibinfo{author}{K.~Li},
  \bibinfo{author}{Y.~Li}, \bibinfo{author}{K.~Lim},
  \bibinfo{author}{B.~Lima~Padovani}, \bibinfo{author}{K.~Lind{\'e}n},
  \bibinfo{author}{N.~Ljube{\v s}i{\'c}}, \bibinfo{author}{O.~Loginova},
  \bibinfo{author}{A.~Luthfi}, \bibinfo{author}{M.~Luukko},
  \bibinfo{author}{O.~Lyashevskaya}, \bibinfo{author}{T.~Lynn},
  \bibinfo{author}{V.~Macketanz}, \bibinfo{author}{A.~Makazhanov},
  \bibinfo{author}{M.~Mandl}, \bibinfo{author}{C.~Manning},
  \bibinfo{author}{R.~Manurung}, \bibinfo{author}{B.~Mar{\c s}an},
  \bibinfo{author}{C.~M{\u a}r{\u a}nduc}, \bibinfo{author}{D.~Mare{\v c}ek},
  \bibinfo{author}{K.~Marheinecke},
  \bibinfo{author}{H.~Mart{\'{\i}}nez~Alonso}, \bibinfo{author}{A.~Martins},
  \bibinfo{author}{J.~Ma{\v s}ek}, \bibinfo{author}{H.~Matsuda},
  \bibinfo{author}{Y.~Matsumoto}, \bibinfo{author}{A.~Mazzei},
  \bibinfo{author}{R.~{McDonald}}, \bibinfo{author}{S.~{McGuinness}},
  \bibinfo{author}{G.~Mendon{\c c}a}, \bibinfo{author}{N.~Miekka},
  \bibinfo{author}{K.~Mischenkova}, \bibinfo{author}{M.~Misirpashayeva},
  \bibinfo{author}{A.~Missil{\"a}}, \bibinfo{author}{C.~Mititelu},
  \bibinfo{author}{M.~Mitrofan}, \bibinfo{author}{Y.~Miyao},
  \bibinfo{author}{A.~Mojiri~Foroushani}, \bibinfo{author}{J.~Moln{\'a}r},
  \bibinfo{author}{A.~Moloodi}, \bibinfo{author}{S.~Montemagni},
  \bibinfo{author}{A.~More}, \bibinfo{author}{L.~Moreno~Romero},
  \bibinfo{author}{G.~Moretti}, \bibinfo{author}{K.~S. Mori},
  \bibinfo{author}{S.~Mori}, \bibinfo{author}{T.~Morioka},
  \bibinfo{author}{S.~Moro}, \bibinfo{author}{B.~Mortensen},
  \bibinfo{author}{B.~Moskalevskyi}, \bibinfo{author}{K.~Muischnek},
  \bibinfo{author}{R.~Munro}, \bibinfo{author}{Y.~Murawaki},
  \bibinfo{author}{K.~M{\"u}{\"u}risep}, \bibinfo{author}{P.~Nainwani},
  \bibinfo{author}{M.~Nakhl{\'e}}, \bibinfo{author}{J.~I.
  Navarro~Hor{\~n}iacek}, \bibinfo{author}{A.~Nedoluzhko},
  \bibinfo{author}{G.~Ne{\v s}pore-B{\=e}rzkalne}, \bibinfo{author}{M.~Nevaci},
  \bibinfo{author}{L.~Nguy{\~{\^e}}n~Th{\d i}},
  \bibinfo{author}{H.~Nguy{\~{\^e}}n Th{\d i}~Minh},
  \bibinfo{author}{Y.~Nikaido}, \bibinfo{author}{V.~Nikolaev},
  \bibinfo{author}{R.~Nitisaroj}, \bibinfo{author}{A.~Nourian},
  \bibinfo{author}{H.~Nurmi}, \bibinfo{author}{S.~Ojala},
  \bibinfo{author}{A.~K. Ojha}, \bibinfo{author}{A.~Ol{\'u}{\`o}kun},
  \bibinfo{author}{M.~Omura}, \bibinfo{author}{E.~Onwuegbuzia},
  \bibinfo{author}{P.~Osenova}, \bibinfo{author}{R.~{\"O}stling},
  \bibinfo{author}{L.~{\O}vrelid}, \bibinfo{author}{{\c S}.~B. {\"O}zate{\c
  s}}, \bibinfo{author}{M.~{\"O}z{\c c}elik},
  \bibinfo{author}{A.~{\"O}zg{\"u}r}, \bibinfo{author}{B.~{\"O}zt{\"u}rk~Ba{\c
  s}aran}, \bibinfo{author}{H.~H. Park}, \bibinfo{author}{N.~Partanen},
  \bibinfo{author}{E.~Pascual}, \bibinfo{author}{M.~Passarotti},
  \bibinfo{author}{A.~Patejuk}, \bibinfo{author}{G.~Paulino-Passos},
  \bibinfo{author}{A.~Peljak-{\L}api{\'n}ska}, \bibinfo{author}{S.~Peng},
  \bibinfo{author}{C.-A. Perez}, \bibinfo{author}{N.~Perkova},
  \bibinfo{author}{G.~Perrier}, \bibinfo{author}{S.~Petrov},
  \bibinfo{author}{D.~Petrova}, \bibinfo{author}{J.~Phelan},
  \bibinfo{author}{J.~Piitulainen}, \bibinfo{author}{T.~A. Pirinen},
  \bibinfo{author}{E.~Pitler}, \bibinfo{author}{B.~Plank},
  \bibinfo{author}{T.~Poibeau}, \bibinfo{author}{L.~Ponomareva},
  \bibinfo{author}{M.~Popel}, \bibinfo{author}{L.~Pretkalni{\c n}a},
  \bibinfo{author}{S.~Pr{\'e}vost}, \bibinfo{author}{P.~Prokopidis},
  \bibinfo{author}{A.~Przepi{\'o}rkowski}, \bibinfo{author}{T.~Puolakainen},
  \bibinfo{author}{S.~Pyysalo}, \bibinfo{author}{P.~Qi},
  \bibinfo{author}{A.~R{\"a}{\"a}bis}, \bibinfo{author}{A.~Rademaker},
  \bibinfo{author}{T.~Rama}, \bibinfo{author}{L.~Ramasamy},
  \bibinfo{author}{C.~Ramisch}, \bibinfo{author}{F.~Rashel},
  \bibinfo{author}{M.~S. Rasooli}, \bibinfo{author}{V.~Ravishankar},
  \bibinfo{author}{L.~Real}, \bibinfo{author}{P.~Rebeja},
  \bibinfo{author}{S.~Reddy}, \bibinfo{author}{G.~Rehm},
  \bibinfo{author}{I.~Riabov}, \bibinfo{author}{M.~Rie{\ss}ler},
  \bibinfo{author}{E.~Rimkut{\.e}}, \bibinfo{author}{L.~Rinaldi},
  \bibinfo{author}{L.~Rituma}, \bibinfo{author}{L.~Rocha},
  \bibinfo{author}{E.~R{\"o}gnvaldsson}, \bibinfo{author}{M.~Romanenko},
  \bibinfo{author}{R.~Rosa}, \bibinfo{author}{V.~Roșca},
  \bibinfo{author}{D.~Rovati}, \bibinfo{author}{O.~Rudina},
  \bibinfo{author}{J.~Rueter}, \bibinfo{author}{K.~R{\'u}narsson},
  \bibinfo{author}{S.~Sadde}, \bibinfo{author}{P.~Safari},
  \bibinfo{author}{B.~Sagot}, \bibinfo{author}{A.~Sahala},
  \bibinfo{author}{S.~Saleh}, \bibinfo{author}{A.~Salomoni},
  \bibinfo{author}{T.~Samard{\v z}i{\'c}}, \bibinfo{author}{S.~Samson},
  \bibinfo{author}{M.~Sanguinetti}, \bibinfo{author}{E.~San{\i}yar},
  \bibinfo{author}{D.~S{\"a}rg}, \bibinfo{author}{B.~Saul{\={\i}}te},
  \bibinfo{author}{Y.~Sawanakunanon}, \bibinfo{author}{S.~Saxena},
  \bibinfo{author}{K.~Scannell}, \bibinfo{author}{S.~Scarlata},
  \bibinfo{author}{N.~Schneider}, \bibinfo{author}{S.~Schuster},
  \bibinfo{author}{L.~Schwartz}, \bibinfo{author}{D.~Seddah},
  \bibinfo{author}{W.~Seeker}, \bibinfo{author}{M.~Seraji},
  \bibinfo{author}{M.~Shen}, \bibinfo{author}{A.~Shimada},
  \bibinfo{author}{H.~Shirasu}, \bibinfo{author}{Y.~Shishkina},
  \bibinfo{author}{M.~Shohibussirri}, \bibinfo{author}{D.~Sichinava},
  \bibinfo{author}{J.~Siewert}, \bibinfo{author}{E.~F. Sigurðsson},
  \bibinfo{author}{A.~Silveira}, \bibinfo{author}{N.~Silveira},
  \bibinfo{author}{M.~Simi}, \bibinfo{author}{R.~Simionescu},
  \bibinfo{author}{K.~Simk{\'o}}, \bibinfo{author}{M.~{\v S}imkov{\'a}},
  \bibinfo{author}{K.~Simov}, \bibinfo{author}{M.~Skachedubova},
  \bibinfo{author}{A.~Smith}, \bibinfo{author}{I.~Soares-Bastos},
  \bibinfo{author}{C.~Spadine}, \bibinfo{author}{R.~Sprugnoli},
  \bibinfo{author}{S.~Steingr{\'{\i}}msson}, \bibinfo{author}{A.~Stella},
  \bibinfo{author}{M.~Straka}, \bibinfo{author}{E.~Strickland},
  \bibinfo{author}{J.~Strnadov{\'a}}, \bibinfo{author}{A.~Suhr},
  \bibinfo{author}{Y.~L. Sulestio}, \bibinfo{author}{U.~Sulubacak},
  \bibinfo{author}{S.~Suzuki}, \bibinfo{author}{Z.~Sz{\'a}nt{\'o}},
  \bibinfo{author}{D.~Taji}, \bibinfo{author}{Y.~Takahashi},
  \bibinfo{author}{F.~Tamburini}, \bibinfo{author}{M.~A.~C. Tan},
  \bibinfo{author}{T.~Tanaka}, \bibinfo{author}{S.~Tella},
  \bibinfo{author}{I.~Tellier}, \bibinfo{author}{M.~Testori},
  \bibinfo{author}{G.~Thomas}, \bibinfo{author}{L.~Torga},
  \bibinfo{author}{M.~Toska}, \bibinfo{author}{T.~Trosterud},
  \bibinfo{author}{A.~Trukhina}, \bibinfo{author}{R.~Tsarfaty},
  \bibinfo{author}{U.~T{\"u}rk}, \bibinfo{author}{F.~Tyers},
  \bibinfo{author}{S.~Uematsu}, \bibinfo{author}{R.~Untilov},
  \bibinfo{author}{Z.~Ure{\v s}ov{\'a}}, \bibinfo{author}{L.~Uria},
  \bibinfo{author}{H.~Uszkoreit}, \bibinfo{author}{A.~Utka},
  \bibinfo{author}{S.~Vajjala}, \bibinfo{author}{R.~van~der Goot},
  \bibinfo{author}{M.~Vanhove}, \bibinfo{author}{D.~van Niekerk},
  \bibinfo{author}{G.~van Noord}, \bibinfo{author}{V.~Varga},
  \bibinfo{author}{E.~Villemonte de~la Clergerie}, \bibinfo{author}{V.~Vincze},
  \bibinfo{author}{N.~Vlasova}, \bibinfo{author}{A.~Wakasa},
  \bibinfo{author}{J.~C. Wallenberg}, \bibinfo{author}{L.~Wallin},
  \bibinfo{author}{A.~Walsh}, \bibinfo{author}{J.~X. Wang},
  \bibinfo{author}{J.~N. Washington}, \bibinfo{author}{M.~Wendt},
  \bibinfo{author}{P.~Widmer}, \bibinfo{author}{S.~Williams},
  \bibinfo{author}{M.~Wir{\'e}n}, \bibinfo{author}{C.~Wittern},
  \bibinfo{author}{T.~Woldemariam}, \bibinfo{author}{T.-s. Wong},
  \bibinfo{author}{A.~Wr{\'o}blewska}, \bibinfo{author}{M.~Yako},
  \bibinfo{author}{K.~Yamashita}, \bibinfo{author}{N.~Yamazaki},
  \bibinfo{author}{C.~Yan}, \bibinfo{author}{K.~Yasuoka},
  \bibinfo{author}{M.~M. Yavrumyan}, \bibinfo{author}{A.~B. Yenice},
  \bibinfo{author}{O.~T. Y{\i}ld{\i}z}, \bibinfo{author}{Z.~Yu},
  \bibinfo{author}{Z.~{\v Z}abokrtsk{\'y}}, \bibinfo{author}{S.~Zahra},
  \bibinfo{author}{A.~Zeldes}, \bibinfo{author}{H.~Zhu},
  \bibinfo{author}{A.~Zhuravleva}, \bibinfo{author}{R.~Ziane},
  \bibinfo{title}{Universal dependencies 2.8.1}, \bibinfo{year}{2021}.
  \bibinfo{note}{{LINDAT}/{CLARIAH}-{CZ} digital library at the Institute of
  Formal and Applied Linguistics ({{\'U}FAL}), Faculty of Mathematics and
  Physics, Charles University}.
\bibitem[{Kondratyuk and Straka(2019)}]{kondratyuk201975}
\bibinfo{author}{D.~Kondratyuk}, \bibinfo{author}{M.~Straka},
\newblock \bibinfo{title}{75 languages, 1 model: Parsing universal dependencies
  universally},
\newblock in: \bibinfo{booktitle}{Proceedings of the 2019 Conference on
  Empirical Methods in Natural Language Processing and the 9th International
  Joint Conference on Natural Language Processing (EMNLP-IJCNLP)},
  \bibinfo{year}{2019}, pp. \bibinfo{pages}{2779--2795}.
\bibitem[{Wolf et~al.(2020)Wolf, Debut, Sanh, Chaumond, Delangue, Moi, Cistac,
  Rault, Louf, Funtowicz, Davison, Shleifer, von Platen, Ma, Jernite, Plu, Xu,
  Le~Scao, Gugger, Drame, Lhoest, and Rush}]{wolf-etal-2020-transformers}
\bibinfo{author}{T.~Wolf}, \bibinfo{author}{L.~Debut},
  \bibinfo{author}{V.~Sanh}, \bibinfo{author}{J.~Chaumond},
  \bibinfo{author}{C.~Delangue}, \bibinfo{author}{A.~Moi},
  \bibinfo{author}{P.~Cistac}, \bibinfo{author}{T.~Rault},
  \bibinfo{author}{R.~Louf}, \bibinfo{author}{M.~Funtowicz},
  \bibinfo{author}{J.~Davison}, \bibinfo{author}{S.~Shleifer},
  \bibinfo{author}{P.~von Platen}, \bibinfo{author}{C.~Ma},
  \bibinfo{author}{Y.~Jernite}, \bibinfo{author}{J.~Plu},
  \bibinfo{author}{C.~Xu}, \bibinfo{author}{T.~Le~Scao},
  \bibinfo{author}{S.~Gugger}, \bibinfo{author}{M.~Drame},
  \bibinfo{author}{Q.~Lhoest}, \bibinfo{author}{A.~Rush},
\newblock \bibinfo{title}{Transformers: State-of-the-art natural language
  processing},
\newblock in: \bibinfo{booktitle}{Proceedings of the 2020 Conference on
  Empirical Methods in Natural Language Processing: System Demonstrations},
  \bibinfo{publisher}{Association for Computational Linguistics},
  \bibinfo{address}{Online}, \bibinfo{year}{2020}, pp. \bibinfo{pages}{38--45}.
  \URLprefix \url{https://aclanthology.org/2020.emnlp-demos.6}.
  \DOIprefix\doi{10.18653/v1/2020.emnlp-demos.6}.
\bibitem[{Devlin et~al.(2019)Devlin, Chang, Lee, and
  Toutanova}]{devlin2019bert}
\bibinfo{author}{J.~Devlin}, \bibinfo{author}{M.-W. Chang},
  \bibinfo{author}{K.~Lee}, \bibinfo{author}{K.~Toutanova},
\newblock \bibinfo{title}{{BERT}: Pre-training of deep bidirectional
  transformers for language understanding},
\newblock in: \bibinfo{booktitle}{Proceedings of the 2019 Conference of the
  North {A}merican Chapter of the Association for Computational Linguistics:
  Human Language Technologies, Volume 1 (Long and Short Papers)},
  \bibinfo{publisher}{Association for Computational Linguistics},
  \bibinfo{address}{Minneapolis, Minnesota}, \bibinfo{year}{2019}, pp.
  \bibinfo{pages}{4171--4186}. \URLprefix
  \url{https://aclanthology.org/N19-1423}.
  \DOIprefix\doi{10.18653/v1/N19-1423}.
\bibitem[{Litschko et~al.(2022)Litschko, Vuli{\'c}, Ponzetto, and
  Glava{\v{s}}}]{litschko2022cross}
\bibinfo{author}{R.~Litschko}, \bibinfo{author}{I.~Vuli{\'c}},
  \bibinfo{author}{S.~P. Ponzetto}, \bibinfo{author}{G.~Glava{\v{s}}},
\newblock \bibinfo{title}{On cross-lingual retrieval with multilingual text
  encoders},
\newblock \bibinfo{journal}{Information Retrieval Journal}
  (\bibinfo{year}{2022}) \bibinfo{pages}{1--35}.
\bibitem[{van~der Goot et~al.(2021)van~der Goot, Sharaf, Imankulova,
  {\"U}st{\"u}n, Stepanovic, Ramponi, Khairunnisa, Komachi, and
  Plank}]{van-der-goot-etal-2020-cross}
\bibinfo{author}{R.~van~der Goot}, \bibinfo{author}{I.~Sharaf},
  \bibinfo{author}{A.~Imankulova}, \bibinfo{author}{A.~{\"U}st{\"u}n},
  \bibinfo{author}{M.~Stepanovic}, \bibinfo{author}{A.~Ramponi},
  \bibinfo{author}{S.~O. Khairunnisa}, \bibinfo{author}{M.~Komachi},
  \bibinfo{author}{B.~Plank},
\newblock \bibinfo{title}{From masked-language modeling to translation:
  Non-{E}nglish auxiliary tasks improve zero-shot spoken language
  understanding},
\newblock in: \bibinfo{booktitle}{Proceedings of the 2021 Conference of the
  North {A}merican Chapter of the Association for Computational Linguistics:
  Human Language Technologies, Volume 1 (Long and Short Papers)},
  \bibinfo{publisher}{Association for Computational Linguistics},
  \bibinfo{address}{Mexico City, Mexico}, \bibinfo{year}{2021}.

\end{thebibliography}

\appendix
\clearpage

\begin{table*}[t]
    \centering
    \resizebox{\linewidth}{!}{
    \begin{tabular}{c|l|rr|rr}
    \toprule
    & \textbf{Dataset} $\rightarrow$ & \multicolumn{2}{c|}{\textsc{Sayfullina}} & \multicolumn{2}{c}{\textsc{SkillSpan}} \\
    \midrule
     & $\downarrow$ \textbf{Method}, \textbf{Metric} $\rightarrow$ & \textbf{Strict} (P $|$ R $|$ F1) & \textbf{Loose} (P $|$ R $|$ F1) & \textbf{Strict} (P $|$ R $|$ F1) & \textbf{Loose} (P $|$ R $|$ F1) \\
     \midrule
\multirow{3}{*}{\rotatebox[origin=c]{90}{\footnotesize\textbf{Baseline}}}
    & Exact         & 9.27 $|$ 1.30 $|$ 2.28    & 25.48 $|$ 3.95 $|$ 6.84     & 23.82 $|$ 3.21 $|$ 5.62         & 43.68 $|$ 8.27 $|$ 13.79  \\
    & Lemmatized    & 8.49 $|$ 1.19 $|$ 2.09    & 25.87 $|$ 4.00 $|$ 6.93     & 23.90 $|$ 2.97 $|$ 5.21         & 41.09 $|$ 7.49 $|$ 12.52 \\
    & POS           & 5.99 $|$ 5.95 $|$ 5.97    & 36.55 $|$ 34.51 $|$ 35.50   & 5.97 $|$ 7.88 $|$ 6.79          & 19.34 $|$ 34.71 $|$ 24.80    \\
    \midrule
\multirow{3}{*}{\rotatebox[origin=c]{90}{\footnotesize\textbf{RoBERTa}}}
    & ISO   & 6.26 $|$ 6.25 $|$ 6.26        & 26.90 $|$ 28.98 $|$ 27.90       & 2.90 $|$ 4.24 $|$ 3.43      & 12.69 $|$ 28.61 $|$ 17.56\\ 
    & AOC   & 3.24 $|$ 3.24 $|$ 3.24        & 64.04 $|$ 55.53 $|$ 59.48       & 2.23 $|$ 2.93 $|$ 2.53      & 20.08 $|$ 37.56 $|$ 26.10\\
    & WSE   & 3.67 $|$ 3.67 $|$ 3.67        & 64.64 $|$ 55.32 $|$ 59.61       & 2.29 $|$ 2.93 $|$ 2.57      & 20.90 $|$ 37.79 $|$ 26.85\\
    \midrule
\multirow{3}{*}{\rotatebox[origin=c]{90}{\footnotesize\textbf{JobBERT}}}
    & ISO   & 7.71 $|$ 7.72 $|$ 7.71        & 27.76 $|$ 29.95 $|$ 28.82                         & 4.17 $|$ 4.65 $|$ 4.39    & 17.07 $|$ 29.48 $|$ 21.61 \\ 
    & AOC   & 4.04 $|$ 4.05 $|$ 4.05        & 56.50 $|$ 48.41 $|$ 52.14                         & 4.44 $|$ 2.96 $|$ 3.54    & 33.64 $|$ 31.28 $|$ 32.30 \\
    & WSE   & 4.15 $|$ 4.16 $|$ 4.15        & 56.98 $|$ 49.00 $|$ 52.69                         & 4.78 $|$ 3.08 $|$ 3.74    & 34.01 $|$ 30.33 $|$ 31.95 \\
    \bottomrule
    \end{tabular}}
    \caption{We show the exact numbers of the performance of the methods.}
    \label{tab:exact}
\end{table*}

\begin{figure*}[t]
    \centering
    \includegraphics[width=\linewidth]{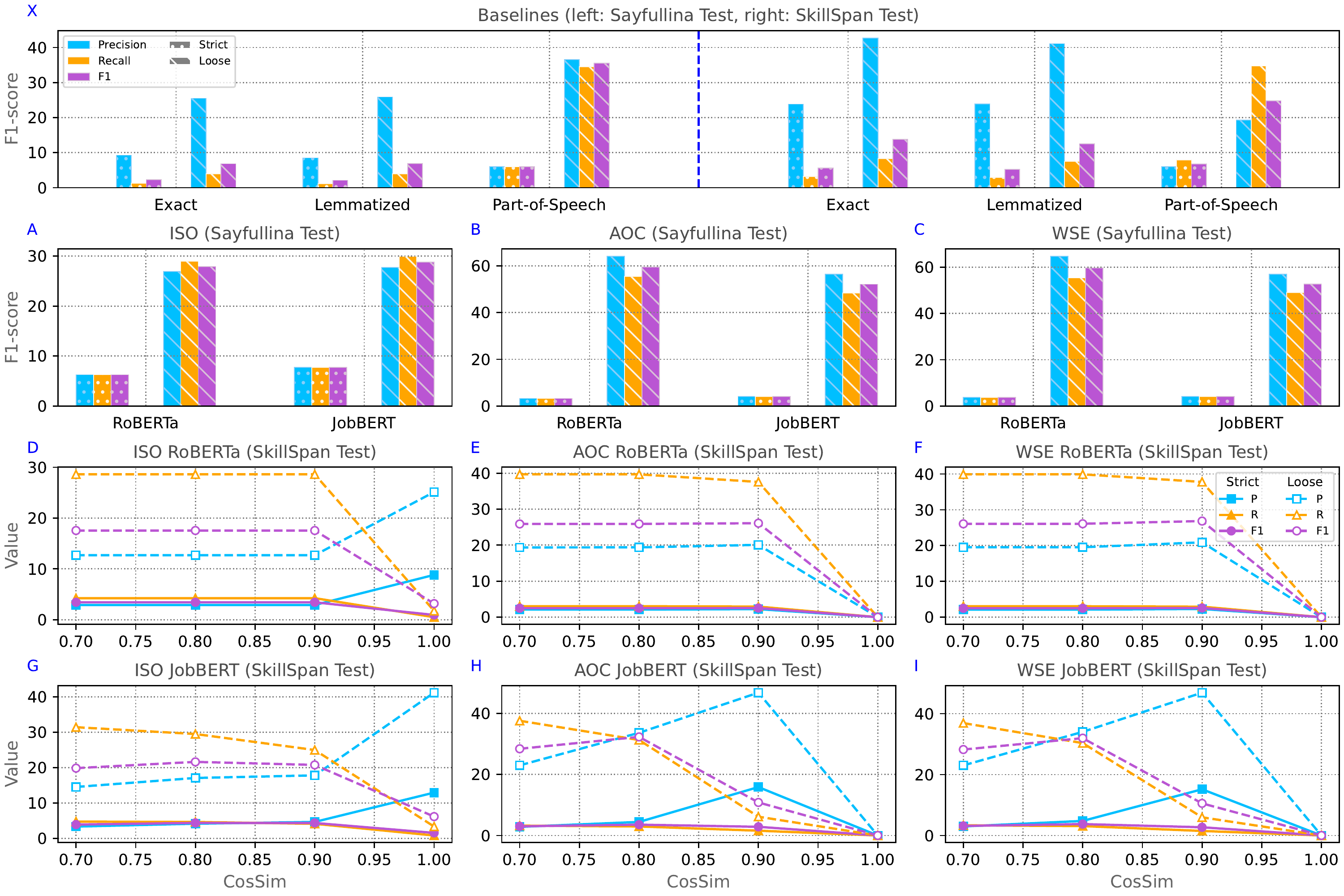}
    \looseness=-1
    \caption{\textbf{Results of Methods.} Results of the baselines are in (X), the performance of ISO, AOC, and WSE on \say{} in (A-C), and the same performance on \sks{} in (D-I) based on the model (RoBERTa or JobBERT). In D--F, we show the precision (P), recall (R), and F1 differences when taking an increasing CosSim.} 
    \label{fig:results}
\end{figure*}

\section{Exact Results}\label{sec:appendix-exact-num}

\paragraph{Definition F1} As mentioned, we evaluate with two types of F1-scores, following~\citet{van-der-goot-etal-2020-cross}. The first type is the commonly used span-F1, where only the correct span and label are counted towards true positives. This is called \texttt{strict-F1}. In the second variant, we seek for partial matches, i.e., overlap between the predicted and gold span including the correct label, which counts towards true positives for precision and recall. This is called \texttt{loose-F1}. We consider the loose variant as well, because we want to analyze whether the span is ``almost correct''.

\paragraph{Exact Numbers Results} We show the exact numbers of \cref{fig:results-f1} in \cref{tab:exact} and more detailed results in~\cref{fig:results}. Results show that there is high precision among the baseline approaches compared to recall. This is balanced using the representation methods for~\say{}. However, we observe that there is much higher recall for \sks{} than precision.

\end{document}